\documentclass[sigconf]{acmart}

\usepackage{times}  
\usepackage{helvet}  
\usepackage{courier}  
\usepackage{url}  
\usepackage{graphicx}  

\usepackage{amsfonts}       
\usepackage{amsmath}
\usepackage{amssymb}
\usepackage{bbm}
\usepackage{amsthm}
\usepackage{bm}
\usepackage{booktabs}       
\usepackage[T1]{fontenc}    
\usepackage{hyperref}       
\usepackage[utf8]{inputenc} 
\usepackage{microtype}      
\usepackage{multirow}
\usepackage{nicefrac}       
\usepackage{subcaption}

\usepackage[english]{babel}
\usepackage{natbib}
\usepackage{lipsum}

\begin{document}
\title{Transfer Meets Hybrid: A Synthetic Approach for Cross-Domain Collaborative Filtering with Text}
\author{Guangneng Hu, Yu Zhang, and Qiang Yang}
\affiliation{Department of Computer Science and Engineering, Hong Kong University of Science and Technology}
\email{njuhgn@gmail.com, yu.zhang.ust@gmail.com, qyang@cse.ust.hk}

\begin{abstract}
Collaborative filtering (CF) is the key technique for recommender systems (RSs). CF exploits user-item behavior interactions (e.g., clicks) only and hence suffers from the data sparsity issue. One research thread is to integrate auxiliary information such as product reviews and news titles, leading to hybrid filtering methods. Another thread is to transfer knowledge from other source domains such as improving the movie recommendation with the knowledge from the book domain, leading to transfer learning methods. In real-world life, no single service can satisfy a user's all information needs. Thus it motivates us to exploit both auxiliary and source information for RSs in this paper. We propose a novel neural model to smoothly enable Transfer Meeting Hybrid (TMH) methods for cross-domain recommendation with unstructured text in an end-to-end manner. TMH attentively extracts useful content from unstructured text via a memory module and selectively transfers knowledge from a source domain via a transfer network. On two real-world datasets, TMH shows better performance in terms of three ranking metrics by comparing with various baselines. We conduct thorough analyses to understand how the text content and transferred knowledge help the proposed model.
\end{abstract}


\maketitle

\section{Introduction}

Recommender systems are widely used in various domains and e-commerce platforms,
such as to help consumers buy products at Amazon, watch videos on Youtube, and read articles on Google News. Collaborative filtering (CF) is among the most effective approaches based on the simple intuition that if users rated items similarly in the past then they are likely to rate items similarly in the future. Matrix factorization (MF) techniques which can learn the latent factors for users and items are its main cornerstone~\cite{PMF,MF}. Recently, neural networks like multilayer perceptron (MLP) are used to learn the interaction function from data~\cite{dziugaite2015neural,he2017neural}. MF and neural CF suffer from the data sparsity and cold-start issues.

One solution is to integrate CF with the content information, leading to hybrid methods. Items are usually associated with content information such as unstructured text, like the news articles and product reviews. These additional sources of information can alleviate the sparsity issue and are essential for recommendation beyond user-item interaction data. For application domains like recommending research papers and news articles, the unstructured text associated with the item is its text content~\cite{CTR,bansal2016ask}. Other domains like recommending products, the unstructured text associated with the item is its user reviews which justify the rating behavior of consumers~\cite{HFT,zheng2017joint}. Topic modelling and neural networks have been proposed to exploit the item content and lead to performance improvement. Memory networks are widely used in question answering and reading comprehension to perform reasoning~\cite{sukhbaatar2015end}. The memories can be naturally used to model additional sources like the item content~\cite{LCMR}, or to model a user's neighborhood who consume the common items with this user~\cite{CMN-Ebesu-18}.

Another solution is to transfer the knowledge from relevant domains and the cross-domain recommendation techniques address such problems~\cite{li2009can,pan2011transfer,cantador2015cross}. In real life, a user typically participates several systems to acquire different information services. For example, a user installs applications in an app store and reads news from a website at the same time. It brings us an opportunity to improve the recommendation performance in the target service (or all services) by learning across domains. Following the above example, we can represent the app installation feedback using a binary matrix where the entries indicate whether a user has installed an app. Similarly, we use another binary matrix to indicate whether a user has read a news article. Typically these two matrices are highly sparse, and it is beneficial to learn them simultaneously. This idea is sharpened into the collective matrix factorization (CMF)~\cite{CMF} approach which jointly factorizes these two matrices by sharing the user latent factors. It combines CF on a target domain and another CF on an auxiliary domain, enabling knowledge transfer~\cite{pan2010survey,zhang2017survey}. In terms of neural networks, given two activation maps from two tasks, cross-stitch convolutional network (CSN)~\cite{CSN} and its sparse variant~\cite{CoNet} learn linear combinations of both the input activations and feed these combinations as input to the successive layers' filters, and hence enabling the knowledge transfer between two domains.

These two threads motivate us to exploit information from both the content and cross-domain information for RSs in this paper. To capture text content and to transfer cross-domain knowledge, we propose a novel neural model, TMH, for cross-domain recommendation with unstructured text in an end-to-end manner. TMH can attentively extract useful content via a memory network (MNet) and can selectively transfer knowledge across domains by a transfer network (TNet), a novel network. A shared layer of feature interactions is stacked on the top to couple the high-level representations learned from individual networks. On real-world datasets, TMH shows the  better performance in terms of ranking metrics by comparing with various baselines. We conduct thorough analyses to understand how the content and transferred knowledge help TMH.

To the best of our knowledge, TMH is the first deep model that transfers cross-domain knowledge for recommendation with unstructured text in an end-to-end learning. Our contributions are summarized as follows:
\begin{itemize}
\item The proposed TMH exploits the text content and transfers the source domain using an attention mechanism which is trained in an end-to-end manner. It is the first deep model that transfers cross-domain knowledge for recommendation with unstructured text using attention based neural networks.
\item We interpret the memory networks to attentively exploit the text content to match word semantics with user preferences. It is among a few recent works on adopting memory networks for hybrid recommendations.
\item The transfer component can selectively transfer source items with the guidance of target user-item interactions by the attentive weights. It is a novel transfer network for cross-domain recommendation.
\item The proposed model can alleviate the sparsity issue including cold-user and cold-item start, and outperforms various baselines in terms of ranking metrics on two real-world datasets.
\end{itemize}

The paper is organized as follows. We firstly introduce the problem formulation in Section~\ref{paper:formulation}. Then we present the memory component to exploit the text content in Section~\ref{paper:MNet} and the transfer component to transfer cross-domain knowledge in Section~\ref{paper:TNet} respectively. We propose the neural model TMH for cross-domain recommendation with unstructured text in Section~\ref{paper:TMH}, followed by its model learning (Section~\ref{paper:learning}) and complexity analyses (Section~\ref{paper:complexity}). In Section~\ref{paper:exp}, we experimentally demonstrate the superior performance of the proposed model over various baselines (Sec.~\ref{exp:results}). We show the benefit of both transferred knowledge and text content for the proposed model in Section~\ref{exp:impact}. We can reduce the amount of cold users and cold items that are difficult to predict accurately (Section~\ref{exp:cold}) and hence alleviate the cold-start issues. We review related works in Section~\ref{related-work} and conclude the paper in Section~\ref{paper:conclusion}.

\section{Related Works}\label{related-work}

\noindent
{\bf Collaborative filtering} Recommender systems aim at learning user preferences on unknown items from their past history. Content-based recommendations are based on the matching between user profiles and item descriptions. It is difficult to build the profile for each user when there is no/few content. Collaborative filtering (CF) alleviates this issue by predicting user preferences based on the user-item interaction behavior, agnostic to the content~\cite{IBCF}. Latent factor models learn feature vectors for users and items mainly based on matrix factorization (MF)~\cite{MF} which has probabilistic interpretations~\cite{PMF}. Factorization machines (FM) can mimic MF~\cite{FM}. To address the data sparsity, an item-item matrix (SPPM) is constructed from the user-item interaction matrix in the CoFactor model~\cite{liang2016factorization}. It then simultaneously factorizes the interaction matrix and the SPPMI matrix in a shared item latent space, enabling the usage of co-click information to regularize the learning of user-item matrix. In contrast with our method, We use independent unstructured text and source domain information to alleviate the data sparsity issue in the user-item matrix.

Neural networks are proposed to push the learning of feature vectors towards non-linear representations, including the neural network matrix factorization (NNMF) and multilayer perceptron (MLP)~\cite{dziugaite2015neural,he2017neural}. The basic MLP architecture is extended to regularize the factors of users and items by social and geographical information~\cite{yang2017bridging}. Other neural approaches learn from the explicit feedback for rating prediction task~\cite{zheng2017joint,catherine2017transnets}. We focus on learning from the implicit feedback for top-N recommendation~\cite{wu2016collaborative}. CF models, however, suffer from the data sparsity issue.

\noindent
{\bf Hybrid filtering} Items are usually associated with the content information such as unstructured text (e.g., abstracts of articles and reviews of products). CF approaches can be extended to exploit the content information~\cite{CTR,CDL,bansal2016ask} and user reviews~\cite{HFT,TBPR,VBPR}. Combining matrix factorization and topic modelling technique (Topic MF) is an effective way to integrate ratings with item contents~\cite{HFT,ling2014ratings,bao2014topicmf}. Item reviews justify the rating behavior of a user, and item ratings are associated with their attributes hidden in reviews~\cite{ganu2009beyond}. Topic MF methods combine item latent factors in ratings with latent topics in reviews~\cite{HFT,bao2014topicmf}. The behavior factors and topic factors are aligned with a link function such as softmax transformation in the hidden factors and hidden topics (HFT) model~~\cite{HFT} or an offset deviation in the collaborative topic regression (CTR) model~~\cite{CTR}. The CTR model assumes the item latent vector learnt from the interaction data is close to the corresponding topic proportions learnt from the text content, but allows them to be divergent from each other if necessary. Additional sources of information are integrated into CF to alleviate the data sparsity issues including knowledge graph~\cite{zhao2017meta,wang2018dkn}. Convolutional networks (CNNs) have been used to extract the features from audio signals for music recommendation~\cite{vandenOord2013DeepCM} and from image for product~\cite{VBPR} and multimedia~\cite{chen2017attentive} recommendation. Autoencoders are used to learn an intermediate representations from text~\cite{CDL,Zhang2016CollaborativeKB}. Recurrent networks~\cite{bansal2016ask} and convolutional networks~\cite{kim2016convolutional,zheng2017joint,catherine2017transnets} can exploit the word order when learning the text representations.

Memory networks can reason with an external memory~\cite{sukhbaatar2015end}. Due to the capability of naturally learning word embeddings to address the problems of word sparseness and semantic gap, a memory module can be used to model item content~\cite{LCMR} or the neighborhood of users~\cite{CMN-Ebesu-18}. Memory networks can learn to match word semantics with the specific user. We follow this research thread by using neural networks to attentively extract important information from the text content.

\noindent
{\bf Cross-domain recommendation} Cross-domain recommendation~\cite{cantador2015cross} is an effective technique to alleviate sparse issue. A class of methods are based on MF applied to each domain. Typical methods include collective matrix factorization (CMF)~\cite{CMF} approach which jointly factorizes two rating matrices by sharing the user latent factors and hence it enables knowledge transfer. CMF has its heterogeneous~\cite{pan2011transfer} variants, and codebook transfer~\cite{li2009can}. The coordinate system transfer can exploit heterogeneous feedbacks~\cite{pan2010transfer,heteroCross}. Multiple source domains~\cite{lu2013selective} and multi-view learning~\cite{MVCross15} are also proposed for integrating information from several domains. Transfer learning (TL) aims at improving the performance of the target domain by exploiting knowledge from source domains~\cite{pan2010survey}. Similar to TL, the multitask learning (MTL) is to leverage useful knowledge in multiple related tasks to help each other~\cite{MTL,zhang2017survey}. The cross-stitch network~\cite{CSN} and its sparse variant~\cite{CoNet} enable information sharing between two base networks for each domain in a deep way. Robust learning is also considered during knowledge transfer~\cite{He2018Robust}. These methods treat knowledge transfer as a global process with shared global parameters and do not match source items with the specific target item given a user. We follow this research thread by using neural networks to selectively transfer knowledge from the source items. We introduce a transfer component to exploit the source domain knowledge.

\section{Problem Formulation}\label{paper:formulation}

For collaborative filtering with implicit feedback, there is a binary matrix $\bm{R} \in \mathbb{R}^{m \times n}$ to describe user-item interactions where each entry $r_{ui} \in \{0,1\}$ is 1 (called observed entries) if user $u$ has an interaction with item $i$ and 0 (unobserved) otherwise:
\[
r_{ui} =\left\{
        \begin{array}{ll}
          1, \quad \textrm{if user-item interaction $(u,i)$ exists}; \\
          0, \quad\textrm{otherwise}.
        \end{array}
      \right.
\]

Denote the set of $m$-sized users by $\mathcal{U}$ and $n$ items by $\mathcal{I}$. Usually the interaction matrix is very sparse since a user $u \in \mathcal{U}$ only consumed a very small subset of all items. Similarly for the task of item recommendation, each user is only interested in identifying top-K items. The items are ranked by their predicted scores:
\begin{equation}\label{eq:function-f}
\hat r_{ui} = f(u,i | \Theta),
\end{equation}
where $f$ is the interaction function and $\Theta$ denotes model parameters.

For MF-based CF approaches, the interaction function $f$ is fixed and computed by a dot product between the user and item vectors. For neural CF, neural networks are used to parameterize function $f$ and learn it from interaction data (see Section~\ref{paper:base}):
\begin{equation}\label{eq:pred-ncf}
f(\bm{x}_{ui} | \bm{P}, \bm{Q}, \theta_f) = \phi_o(...(\phi_1(\bm{x}_{ui}))...),
\end{equation}
where input $\bm{x}_{ui} = [\bm{P}^T\bm{x}_u, \bm{Q}^T\bm{x}_i] \in \mathbb{R}^{2d}$ is concatenated from that of user and item embeddings, which are projections of their one-hot encodings $\bm{x}_u \in \{0,1\}^m$ and $\bm{x}_i \in \{0,1\}^n$ by embedding matrices $\bm{P} \in \mathbb{R}^{m \times d}$ and $\bm{Q} \in \mathbb{R}^{n \times d}$, respectively. The output and hidden layers are computed by $\phi_o$ and $\{\phi_l\}$ in a neural network.

Similarly, for cross-domain recommendation, we have a target domain (e.g., news domain) user-item interaction matrix $\bm{R}_T \in \mathbb{R}^{m \times n_T}$ and a source domain (e.g., app domain) matrix $\bm{R}_S \in \mathbb{R}^{m \times n_S}$ where $m = |\mathcal{U}|$ and $n_T = |\mathcal{I}_T|$ ($n_S = |\mathcal{I}_S|$) is the size of users $\mathcal{U}$ and target items $\mathcal{I}_T$ (source items $\mathcal{I}_S$). Note that the users are shared and hence we can transfer knowledge across domains. We use $u$ to index users, $i$ to target items, and $j$ to source items. Let $[j]^u = (j_1, j_2, ..., j_s)$ be the $s$-sized source items that user $u$ has interacted with in the source domain. Neural CF can be extended to leverage the source domain and then the interaction function has the form of (see Section~\ref{paper:TNet}):
\begin{equation}\label{eq:pred-transfer}
 f(u,i, [j]^u | \Theta).
\end{equation}

Usually, for hybrid filtering, the (target) domain also has the content information (e.g., product reviews). Denote by $d_{ui}$ the content text corresponding to user $u$ and item $i$. It is a sequence of words $d_{ui} = (w_1, w_2, ..., w_l)$ where each word $w$ comes from a vocabulary $\mathcal{V}$ and $l = |d_{ui}|$ is the length of the text document. Neural CF can be extended to leverage text and then the interaction function has the form of  (see Section~\ref{paper:MNet}):
\begin{equation}\label{eq:pred-hybrid}
 f(u,i, d_{ui} | \Theta).
\end{equation}

For the task of item recommendation, the goal is to generate a ranked list of items for each user based on her history records, i.e., top-N recommendations. We hope improve the recommendation performance in the target domain with the help of both the content and source domain information.

A synthetic model of transfer and hybrid estimates the probability of his/her each observation conditioned on this user, the content text, and the interacted source items:
\begin{equation}\label{eq:prob-one-observation}
\hat r_{ui} \triangleq  p(r_{ui} = 1 | u, d_{ui}, [j]^u).
\end{equation}
The equation sharpens the intuition behind the synthetic model, that is, the conditional probability of whether user $u$ will like the item $i$ can be determined by three factors: 1) his/her individual preferences, 2) the corresponding content text ($d_{ui}$), and 3) his/her behavior in a related source domain ($[j]^u$). The likelihood function of the entire matrix $\mathbf{R}_T$ is then defined as:
\begin{equation}\label{eq:likelihood-all}
p(\mathbf{R}_T) = \Pi_u \Pi_i  p(r_{ui} | u, d_{ui}, [j]^u).
\end{equation}

We propose a novel neural model to learn the conditional probability in an end-to-end manner (see Section~\ref{paper:TMH}):
\begin{equation}
\hat r_{ui} = f(u,i, d_{ui}, [j]^u | \Theta_f),
\end{equation}
where $f$ is the network function and $\Theta_f$ are model parameters.

The model consists of a memory network $\mathbf{o}_{ui} = f_M(u,i,d_{ui} | \Theta_M)$ to model unstructured text (Sec.~\ref{paper:MNet}) and a transfer network $\mathbf{c}_{ui} = f_T(i,[j]^u | \Theta_T)$ to transfer knowledge from the source domain (Sec.~\ref{paper:TNet}). A shared feature interaction layer $f_S(\mathbf{o}_{ui}, \mathbf{z}_{ui}, \mathbf{c}_{ui} | \Theta_S)$ where $\mathbf{z}_{ui}$ is the non-linear representations of the $(u,i)$ interaction, is stacked on the top of the learned high-level representations from individual networks (Sec.~\ref{paper:TMH}).

\section{A Basic Neural CF Network}\label{paper:base}

We adopt a feedforward neural network (FFNN) as the base neural CF model to parameterize the interaction function (see Eq. (\ref{eq:pred-ncf})). The basic network is similar to the Deep model in~\cite{Covington2016DeepNN,cheng2016wide} and the MLP model in~\cite{he2017neural}. The base network consists of four modules with the information flow from the input $(u,i)$ to the output $\hat{r}_{ui}$ as follows.

\noindent
{\bf Input: $(u,i) \rightarrow \mathbbm{1}_u,\mathbbm{1}_i$} This module encodes user-item interaction indices. We adopt the one-hot encoding. It takes user $u$ and item $i$, and maps them into one-hot encodings $\mathbbm{1}_u \in \{0,1\}^{m}$ and $\mathbbm{1}_i \in \{0,1\}^{n}$ where only the element corresponding to that index is 1 and all others are 0.

\noindent
{\bf Embedding: $\mathbbm{1}_u,\mathbbm{1}_i \rightarrow \bm{x}_{ui}$} This module firstly embeds one-hot encodings into continuous representations $\bm{x}_u=\bm{P}^T \mathbbm{1}_u$ and $\bm{x}_i=\bm{Q}^T\mathbbm{1}_i$ by embedding matrices $\bm{P}$ and $\bm{Q}$ respectively, and then concatenates them as $\bm{x}_{ui} = [\bm{x}_u, \bm{x}_i],$ to be the input of following building blocks.

\noindent
{\bf Hidden layers}: $\bm{x}_{ui} \rightsquigarrow \bm{z}_{ui}$. This module takes the continuous representations from the embedding module and then transforms through several layers to a final latent representation $\bm{z}_{ui}= (...(\phi_1(\bm{x}_{ui})...)$. This module consists of hidden layers to learn nonlinear interaction between users and items.

\noindent
{\bf Output }: $\bm{z}_{ui} \rightarrow \hat r_{ui}$. This module predicts the score $\hat r_{ui}$ for the given user-item pair based on the representation $\bm{z}_{ui}$ from the last layer of multi-hop module. Since we focus on one-class collaborative filtering, the output is the probability that the input pair is a positive interaction. This can be achieved by a softmax layer: $\hat r_{ui} = \phi_o(\bm{z}_{ui}) = \frac{1}{1 + \exp(-\bm{h}^T \bm{z}_{ui})},$ where $\bm{h}$ is the parameter.

\begin{table}[]
\centering
\caption{Model Parameters of TMH.}
\label{table:parameter}
\resizebox{0.41\textwidth}{!}{
\begin{tabular}{c|c|c}
\hline
Parameter   & Dimension     & Description \\
\hline
$\bm{P}$    & $m \times d $         & User embedding matrix  \\
\hline
$\bm{Q}$    & $n_T \times d $         & Target item embedding matrix  \\
\hline
$\bm{H}$    & $n_S \times d $         & Source item embedding matrix  \\
\hline
$\bm{A}$    & $L \times 2d $         & Internal memory matrix \\
\hline
$\bm{C}$    & $L \times 2d $         & External memory matrix \\
\hline
\multirow{2}{*}{$\bm{W}, \bm{b}$}  & \multirow{2}{*}{$2d \times d$, $d$} & Linear mapping weight and bias \\
 & & for the user-item interaction\\
\hline
\multirow{2}{*}{$\bm{W}_o, \bm{W}_z, \bm{W}_c$}  & \multirow{2}{*}{$d \times d $}  & Linear mapping for outputs \\
 & & of individual networks \\
\hline
$\bm{h}$    & $3d$    & Weight of the shared layer \\
\hline
\end{tabular}
}
\end{table}

\section{The Proposed TMH Model}\label{paper:TMH}

We describe the proposed Transfer Meet Hybrid (TMH) model in this section. TMH models user preferences in the target domain by exploiting the text content and transferring knowledge from a source/auxiliary domain. TMH learns high-level representations for unstructured text and source domain items such that the learned representations can estimate the conditional probability of that whether a user will like an item. This is done with a memory network (Sec.~\ref{paper:MNet}) and a transfer network (Sec.~\ref{paper:TNet}), coupled by the shared embeddings on the bottom and an interaction layer on the top (Sec.~\ref{paper:TMH-model}). The entire network can be trained efficiently to minimize a binary cross-entropy loss by back-propagation (Sec.~\ref{paper:learning}). We begin by describing the recommendation problem and the model formulation before introducing the network architecture.

\subsection{Matching Word Semantics with User Preferences}\label{paper:MNet}

We adapt a memory network (MNet) to integrate unstructured text since it can learn to match word semantics with user preferences. Memory networks have been used in recommendation to model item content~\cite{huang2017mention}, model users' neighborhood~\cite{ebesu2018collaborative}, and learn latent relationships~\cite{Tay2018LatentRM}. The local and centralized memories recommender (LCMR)~\cite{LCMR} uses a local memory module (LMM) to exploit the text content by using MNet. We use memory networks (MNet) to attentively extract important information from the text content via the attention mechanism which can match word semantics with the specific user and determine which words are highly relevant to the user preferences.

MNet is a variant of memory augmented neural network which can learn high-level representations of unstructured text with respect to the given user-item interaction. The attention mechanism inherent in the memory component can determine which words are highly relevant to the user preferences.

The MNet consists of one internal memory matrix $\bf{A} \in \mathbb{R}$$^{L \times 2d}$ where $L$ is the vocabulary size (typically $L = 8,000$ after processing~\cite{CTR}) and $2d$ is the dimension of each memory slot, and one external memory matrix $\bf{C}$ with the same dimensions as $\bf{A}$. The function of the two memory matrices works as follows.

Given a document $d_{ui} = (w_1, w_2, ..., w_l)$ corresponding to the $(u,i)$ interaction, we form the memory slots $\bm{m}_k \in \mathbb{R}^{2d}$ by mapping each word $w_k$ into an embedding vector with matrix $\bf{A}$, where $k=1,...,l$ and the length of the longest document is the memory size. We form a preference vector $\bm{q}^{(ui)}$ corresponding to the given document $d_{ui}$ and the user-item interaction $(u,i)$ where each element encodes the relevance of user $u$ to these words given item $i$ as:
\begin{equation}\label{eq:attention-user-item}
q_k^{(ui)} = \bm{x}_u^T \bm{m}_k^{(u)} + \bm{x}_i^T \bm{m}_k^{(i)}, \; k=1,...,l
\end{equation}
where we split the $\bm{m}_k = [\bm{m}_k^{(u)}, \bm{m}_k^{(i)}]$ into the user part $\bm{m}_k^{(u)}$ and the item part $\bm{m}_k^{(i)}$. The $\bm{x}_u$ and $\bm{x}_i$ are the user and item embeddings obtained by embedding matrices $\bm{P} \in \mathbb{R}$$^{m \times d}$ and $\bm{Q} \in \mathbb{R}$$^{n_T \times d}$ respectively. On the right hand of the above equation, the first term captures the matching between preferences of user $u$ and word semantics, for example, the user is a machine learning researcher and he/she may be more interested in the words such as ``optimization'' and ``Bayesian'' than those of ``history'' and ``philosophy''. The second term computes the support of item $i$ to the words, for example, the item is a machine learning related article and it may support more the words such as ``optimization'' and ``Bayesian'' than those of ``history'' and ``philosophy''. Together, the content-based/associative addressing scheme can determine internal memories with highly relevance to the target user $u$ regarding the words $d_{ui}$ given the specific item $i$.

\begin{figure}
\centering
\includegraphics[height=2.4in,width=6.8cm]{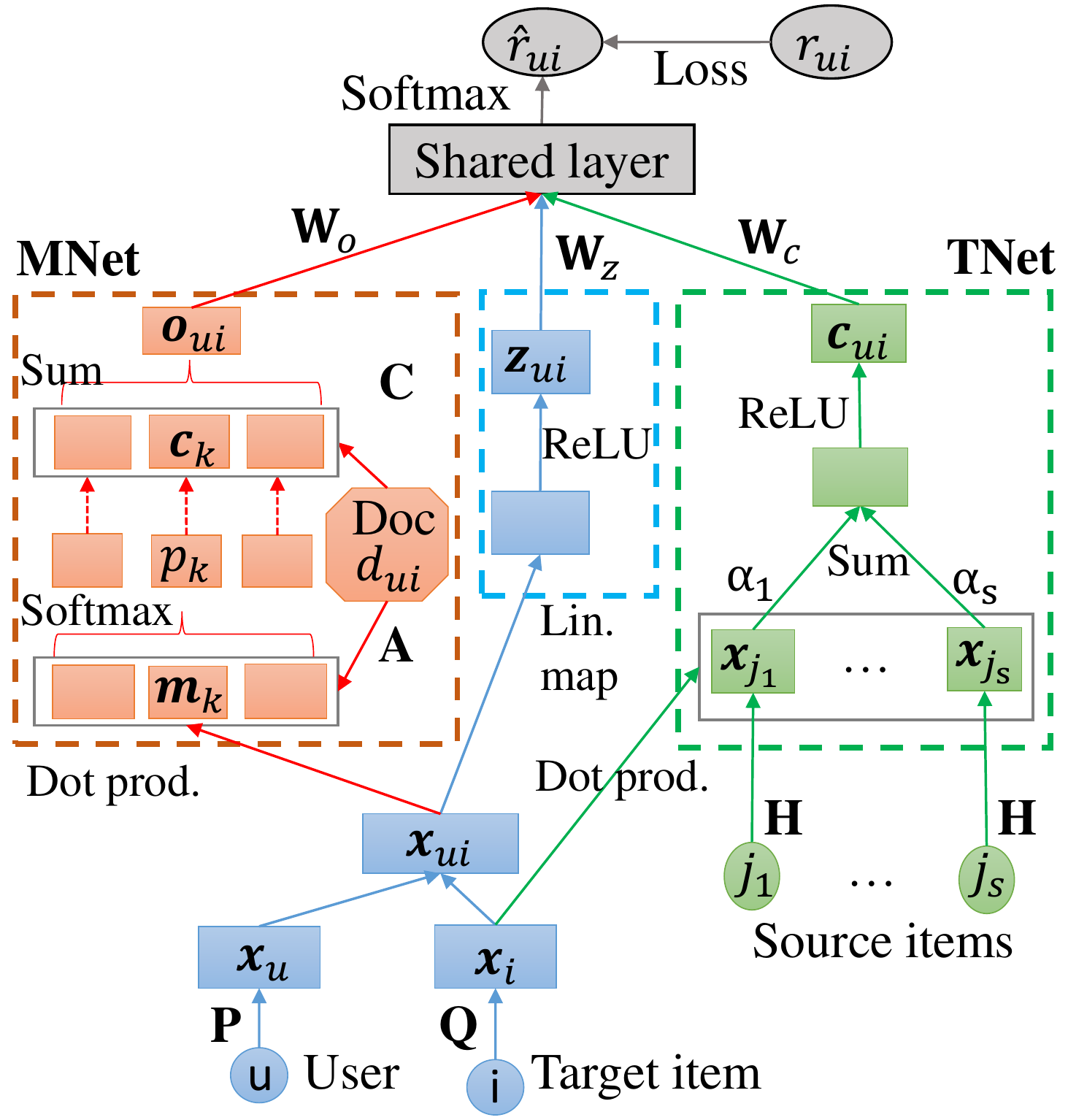}
\caption{The Architecture of TMH. The memory net (MNet) is to model unstructured text. The transfer net (TNet) is to transfer cross-domain knowledge. The CF net is to learn the user-item interaction function. The shared layer is to learn feature interactions from the outputs of three individual networks.}
\label{fig:TMH}
\end{figure}

Actually we can compact the above two terms with a single vector dot product by concatenating the embeddings of the user and the item into $\bm{x}_{ui} = [\bm{x}_u, \bm{x}_i]$: $
q_k^{(ui)} = \bm{x}_{ui}^T \bm{m}_k.$ The neural attention mechanism can adaptively learn the weighting function over the words to focus on a subset of them. Traditional combination of words predefines a heuristic weighting function such as average or weights with tf-idf scores. Instead, we compute the attentive weights over words for a given user-item interaction to infer the importance of each word's unique contribution:
\begin{equation}
p_k^{(ui)} = Softmax(q_k^{(ui)}) = \frac{\exp(\beta q_k^{(ui)})}{\sum_{k'} \exp(\beta q_{k'}^{(ui)})},
\vspace{-1.5mm}
\end{equation}
which produces a probability distribution over the words in $d_{ui}$. The neural attention mechanism allows the memory component to focus on specific words while to place little importance on other words which may be less relevant. The parameter $\beta$ is introduced to stabilize the numerical computation when the exponentials of the softmax function are very large and it also can amplify or attenuate the precision of the attention like a temperature~\cite{hinton2015distilling} where a higher temperature (i.e., a smaller $\beta$) produces a softer probability distribution over words. We set $\beta = d^{-\frac{1}{2}}$ by scaling along with the dimensionality~\cite{vaswani2017attention}.

We construct the high-level representations by interpolating the external memories with the attentive weights as the output:
\begin{equation}\label{eq:weighted-sum}
\bm{o}_{ui} = \sum\nolimits_k p_k^{(ui)} \bm{c}_k,
\end{equation}
where the external memory slot $\bm{c}_k \in \mathbb{R}^d$ is another embedding vector for word $w_k$ by mapping it with matrix $\bf{C}$. The external memories allows the storage of long-term knowledge pertaining specifically to each word's role in matching the user preference. In other words, the content-based addressing scheme identifies important words in a document acting as a key to retrieval the relevant values stored in the external memory matrix $\bm{C}$ via the neural attention mechanism. The attention mechanism adaptively weights words according to the specific user and item. The final output $\bm{o}_{ui}$ represents a high-level, summarized information extracted attentively from the text content involved with relations between the user-item interaction $(u,i)$ and the corresponding words $d_{ui}$.

A more detailed discussion on Eq. (\ref{eq:attention-user-item}) is in order. One alternative to form the memory slot $\bm{m}_k$ is to use the pre-trained word embedding $\bm{e}_k$ for word $w_k$. If the dimensions of $\bm{e}_k$ and $\bm{m}_k$ are different, then a matrix $\bf{A}$ can be used to build the connection between them as:  $\bm{m}_k = \bf{A} \bm{e}_k$. We, however, compute the attentions on words using the embeddings of the corresponding user and item, rather than the pre-trained word embeddings themselves. The reason is as follows. Though some sentimental words like `good' and `bad' are somewhat important, they also depend on who wrote the reviews~\cite{tang2015user}. Some people are critical and hence when they give a `good' word in the reivew, it means that the product is really good. While some people are very kind, they usually give `good' words to all products. We address this issue by taking the user information into account when computing the attention weights, as motivated by~\cite{tang2015user}. Moreover, the sentimental words only exist in review-related corpus, and they do not exist in non-emotional datasets like scientific articles. As a result, our way of computing attentions on words is general and applicable to many settings.

\noindent
{\bf Remark I} MNet attentively extracts useful content to match the word semantics with specific user where different words in the text document have different weights in the semantic factor. Memory networks\cite{Weston2015MemoryN} are firstly proposed to address the question answering (QA) task where memories are a short story and the query is a question related to the text in which the answer can be reasoned by the network. We can think of the recommendation with text as a QA problem: the question to be answered is to ask how likely a user prefers an item. The unstructured text is analogue to the story and the query is analogue to the user-item interaction.

\subsection{Selecting Source Items to Transfer}\label{paper:TNet}

We introduce a transfer component to exploit the source domain knowledge. A user may participate several systems to acquire different information needs, for example, a user installs apps in an app store and reads news from other website. Cross-domain recommendation~\cite{cantador2015cross} is an effective technique to alleviate sparse issue where transfer learning (including multitask learning)~\cite{pan2010survey,MTL,zhang2017survey} is a class of underlying methods. Typical methods include collective matrix factorization (CMF)~\cite{CMF} approach which jointly factorizes two rating matrices by sharing the user latent factors and hence it enables knowledge transfer. The cross-stitch network~\cite{CSN} and its sparse variant~\cite{CoNet} enable information sharing between two base networks for each domain in a deep way. These methods treat knowledge transfer as a global process (shared global parameters) and do not match source items with the specific target item given a user.

We propose a novel transfer network (TNet) which can selectively transfer source knowledge for specific target item. Since the relationships between items are shown to be important in improving recommendation performance~\cite{Paterek2007ImprovingRS,koren2008factorization,ning2011slim,Nguyen2018NPENP} for single domain, {\em we want to capture relationships between target item and source items of a user. The central idea is to learn adaptive weights over source items specific to the given target item during the knowledge transfer.}

Given the source items $[j]^u = (j_1, j_2, ..., j_s)$ with which the user $u$ has interacted in the source domain, TNet learns a transfer vector $\bm{c}_{ui} \in \mathbb{R}^d$ to capture the relations between the target item $i$ and source items given the user $u$. The underlying observations can be illustrated in an example of improving the movie recommendation by transferring knowledge from the book domain. When we predict the preference of a user on the movie ``The Lord of the Rings,'' the importance of her read books such as ``The Hobbit,'' and ``The Silmarillion'' may be much higher than those such as ``Call Me by Your Name''.

The similarities between target item $i$ and source items can be computed by their dot products:
\begin{equation}
a_j^{(i)} = \bm{x}_i^T \bm{x}_j, \; j=1,...,s,
\end{equation}
where $\bm{x}_j \in  \mathbb{R}^d$ is the embedding for the source item $j$ by an embedding matrix $\bm{H} \in \mathbb{R}$$^{n_S \times d}$. This score computes the compatibility between the target item and the source items consumed by the user. For example, the similarity of target movie $i = $ ``The Lord of the Rings,'' with the source book $j = $ ``The Hobbit'' may be larger than that with the source book $j' = $ ``Call Me by Your Name'' (given a user $u$).

We normalize similarity scores to be a probability distribution over source items:
\begin{equation}
\alpha_j^{(i)}= Softmax(a_j^{(i)}),
\end{equation}
and then the transfer vector is a weighted sum of the corresponding source item embeddings:
\begin{equation}\label{eq:cross-sum}
\bm{c}_{ui} = ReLU(\sum\nolimits_j \alpha_j^{(i)} \bm{x}_j),
\end{equation}
where we introduce non-linearity on the transfer vector by activation function rectified linear unit (ReLU). Empirically we found that the activation function $ReLU(x) = max(0, x)$ works well due to its non-saturating nature and suitability for sparse data. The transfer vector $\bm{c}_{ui}$ is a high-level representation, summarizing the knowledge from the source domain as the output of the TNet. TNet can selectively transfer representations from the corresponding embeddings of source items with the guidance of the target user-item interaction.

A more detailed discussion on Eq. (\ref{eq:cross-sum}) is in order. Eq. (\ref{eq:cross-sum}) sharpens the idea that the transfer component can selectively transfer source items with the guidance of target user-item interactions. This is achieved by attentive weights $\alpha_j^{(i)}$ (or $a_j^{(i)})$. When the source item $j$ is highly relevant to the target item $i$ given user $u$, then the knowledge from the source domain is easily flowing into the target domain with a high influence weight. When the source item $j$ is irrelevant to the target item $i$ given user $u$, then the knowledge from source domain is hard to flow into the target domain with a small effect weight. This selection is automatically determined by the transfer component, but this is not easily achieved by the existing cross-domain recommendation techniques like the multitask models such as collective matrix factorization~\cite{CMF} and collaborative cross networks~\cite{CoNet} (a variant of cross-stitch networks~\cite{CSN}) which have multi-objective optimization. Besides, we implicitly use the label information from the source domain when generating the source items for a user, while CMF and CSN explicitly exploit label information by learning to predict the labels. As a result, the transfer component benefits from the source domain knowledge in two-step: selecting instances (source items) to transfer via source domain labels and re-weighting instances with attentive weights.

\noindent
{\bf Remark II } The computational process of MNet and TNet is similar. We firstly compute attentive weights over a collection of objects (words in MNet and items in TNet). Then we summarize the high-level representation as the output (the text representation $\bm{o}_{ui}$ in MNet and the transfer vector $\bm{c}_{ui}$ in TNet), weighted by the attentive probabilities which are computed by a content-based addressing scheme.

 \vspace{-2mm}
\subsection{TMH}\label{paper:TMH-model}

The architecture for the proposed TMH model is illustrated in Figure~\ref{fig:TMH} as a feedforward neural network (FFNN). The input layer specifies embeddings of a user $u$, a target item $i$, and the corresponding source items $[j]^u = (j_1,...,j_s)$. The content text $d_{ui}$ is modelled by the memories in the MNet to produce a high-level representation $\bm{o}_{ui}$. The source items are transferred into the transfer vector $\bm{c}_{ui}$ with the guidance of $(u,i)$ in the TNet. These computational pathes are introduced in the above Sec.~\ref{paper:MNet} and Sec.~\ref{paper:TNet} respectively.

Firstly, we use a simple neural CF model (CFNet) which has one hidden layer to learn a nonlinear representation for the user-item interaction:
\begin{equation}
 \bm{z}_{ui} = ReLU(\bm{W} \bm{x}_{ui} + \bm{b}),
\end{equation}
where $\bm{W}$ and $\bm{b}$ are the weight and bias parameters in the hidden layer. Usually the dimension of  $\bm{z}_{ui}$ is half of that $\bm{x}_{ui}$ in a typical tower-pattern architecture.

The outputs from the three individual networks can be viewed high-level features of the content text, source domain knowledge, and the user-item interaction. They come from different feature space learned by different networks. Thus, we use a shared layer on the top of the all features:
\begin{equation}
\hat r_{ui} = \frac{1}{1 + \exp(-\bm{h}^T \bm{y}_{ui})},
\end{equation}
where $\bm{h}$ is the parameter. And the joint representation:
\begin{equation}
\bm{y}_{ui} = [\bm{W}_o \bm{o}_{ui},\bm{W}_z \bm{z}_{ui},\bm{W}_c \bm{c}_{ui}],
\end{equation}
is concatenated from the linear mapped outputs of individual networks where matrices $\bm{W}_o,\bm{W}_z,\bm{W}_c$ are the corresponding linear mapping transformations..


\subsection{Learning}\label{paper:learning}

Due to the nature of the implicit feedback and the task of item recommendation, the squared loss $(\hat r_{ui} - r_{ui})^2$ may be not suitable since it is usually for rating prediction. Instead, we adopt the binary cross-entropy loss:$\mathcal{L} =  - \sum\nolimits_{(u,i) \in \mathcal{S} } r_{ui} \log{\hat r_{ui}} + (1-r_{ui}) \log(1 - \hat r_{ui} ),$ where the training samples $\mathcal{S} = \bm{R}^+_T \cup \bm{R}^-_T$ are the union of observed target interaction matrix and randomly sampled negative pairs. Usually, $|\bm{R}^+_T| = |\bm{R}^-_T|$ and we do not perform a predefined negative sampling in advance since this can only generate a fixed training set of negative samples. Instead, we generate negative samples during each epoch, enabling diverse and augmented training sets of negative examples to be used.

This objective function has a probabilistic interpretation and is the negative logarithm likelihood of the following likelihood function: $L(\Theta | \mathcal{S} ) = \prod\nolimits_{(u,i) \in \bm{R}^+_T} \hat r_{ui} \prod\nolimits_{(u,i) \in \bm{R}^-_T} (1 - \hat r_{ui}),$ where the model parameters are: $\Theta=\{\bm{P}, \bm{Q}, \bm{H}, \bm{A}, \bm{C}, \bm{W}, \bm{b}, \bm{W}_o, \bm{W}_z, \bm{W}_c, \bm{h}\}.$ Comparing with Eq. (\ref{eq:likelihood-all}), instead of modeling all zero entries (i.e., the whole target matrix $\bm{R}_T$), we learn from only a small subset of such unobserved entries and treat them as negative samples by picking them randomly during each optimization iteration (i.e., the negative sampling technique). The objective function can be optimized by stochastic gradient descent (SGD) and its variants like adaptive moment method (Adam)~\cite{Adam}. The update equations are: $\Theta^{new} \leftarrow \Theta^{old} - \eta {\partial L(\Theta)} / {\partial \Theta},$ where $\eta$ is the learning rate. Typical deep learning library like TensorFlow ({\url{https://www.tensorflow.org}) provides automatic differentiation and hence we omit the gradient equations ${\partial L(\Theta)} / {\partial \Theta}$ which can be computed by chain rule in back-propagation (BP).

\subsection{Complexity Analysis }\label{paper:complexity}

In the model parameters $\Theta$, the embedding matrices $\bm{P}$, $\bm{Q}$ and $\bm{H}$ contain a large number of parameters since they depend on the input size of users and (target and source) items, and their scale is hundreds of thousands. Typically, the number of words, i.e., the vocabulary size is $L = 8,000$~\cite{CTR}. The dimension of embeddings is typically $d=100$. Since the architecture follows a tower pattern, the dimension of the outputs of the three individual networks is also limited within hundreds. In total, the size of model parameters is linear with the input size and is close to the size of typical latent factors models~\cite{CMF} and one hidden layer neural CF approaches~\cite{he2017neural}.

During training, we compute the outputs of the three individual networks in parallel using mini-batch stochastic optimization which can be trained efficiently by back-propagation. TMH is scalable to the number of the training data. It can easily update when new data examples come, just feeding them into the training mini-batch. Thus, TMH can handle the scalability and dynamics of items and users like in an online fashion. In contrast, the topic modeling related techniques have difficulty in benefitting from these advantages to this extent.

\section{Experiments}\label{paper:exp}

In this section, we conduct empirical study to answer the following questions: 1) how does the proposed TMH model perform compared with state-of-the-art recommender systems; and 2) how do the text content and the source domain information contribute each to the proposed framework. We firstly introduce the evaluation protocols and experimental settings, and then we compare the performance of different recommender systems. We further analyze the TMH model to understand the impact of the memory and transfer component. We also investigate that the improved performance comes from the cold-users and cold-items to some extent.

\subsection{Experimental Settings}\label{exp:setting}

\begin{table}[]
\centering
\caption{Datasets and Statistics.}
\label{table:data}
\resizebox{0.42\textwidth}{!}{
\begin{tabular}{c | c | c | r}
\hline \hline
Dataset                 & Domain                  & Statistics &  Amount \\
\hline \hline
\multirow{9}{*}{Mobile News} & Shared             & \#Users    &  15,890 \\
\cline{2-4}
                        & \multirow{5}{*}{Target} & \#News     &  84,802 \\
                        &                         & \#Reads    &  477,685 \\
                        &                         & Density    &  0.035\% \\
\cline{3-4}
                        &                         & \#Words    &  612,839 \\
                        &                         & Avg. Words Per News   & 7.2 \\
\cline{2-4}
                        & \multirow{3}{*}{Source} & \#Apps     & 14,340 \\
                        &                         & \#Installations       & 817,120 \\
                        &                         & Density    &  0.359\% \\
\hline \hline
\multirow{9}{*}{Amazon Product} &  Shared             & \#Users    & 8,514   \\
\cline{2-4}
                        & \multirow{5}{*}{Target} & \#Clothes (Men)       & 28,262  \\
                        &                         & \#Ratings/\#Reviews   & 56,050   \\
                        &                         & Density    & 0.023\%   \\
\cline{3-4}
                        &                         & \#Words    &   1,845,387 \\
                        &                         & Avg. Words Per Review & 32.9  \\
\cline{2-4}
                        & \multirow{3}{*}{Source} & \#Products (Sports)   & 41,317 \\
                        &                         & \#Ratings/\#Reviews   & 81,924 \\
                        &                         & Density    &  0.023\% \\
\hline \hline
\end{tabular}
}
\end{table}

\noindent
{\bf Dataset } We evaluate on two real-world cross-domain datasets. The first dataset, {\bf Mobile}\footnote{An anonymous version can be released later.}, is provided by a large internet company, i.e., Cheetah Mobile ({\url{http://www.cmcm.com/en-us/})~\cite{liu2018transferable}. The information contains logs of user reading news, the history of app installation, and some metadata such as news publisher and user gender collected in one month in the US. We removed users with fewer than 10 feedbacks. For each item, we use the news title as its text content. Following the work \cite{CTR}, we filter stop words and use tf-idf to choose the top 8,000 distinct words as the vocabulary. This yields a corpus of 612K words. The average number of words per news is less than 10. The dataset we used contains 477K user-news reading records and 817K user-app installations. There are 15.8K shared users which enable the knowledge transfer between the two domains. We aim to improve the news recommendation by transferring knowledge from app domain. The data sparsity is over 99.6\%.

The second dataset is a public {\bf Amazon} dataset ({\url{http://snap.stanford.edu/data/web-Amazon.html}), which has been widely used to evaluate the performance of collaborative filtering approaches~\cite{VBPR}. We use the two categories of Amazon Men and Amazon Sports as the cross-domain~\cite{VBPR,TBPR}. The original ratings are from 1 to 5 where five stars indicate that the user shows a positive preference on the item while the one stars are not. We convert the ratings of 4-5 as positive samples. The dataset we used contains 56K positive ratings on Amazon Men and 81K positive ratings on Amazon Sports. There are 8.5K shared users, 28K Men products, and 41K Sports goods. We aim to improve the recommendation on the Men domain by transferring knowledge from relevant Sports domain. The data sparsity is over 99.7\%. We filter stop words and use tf-idf to choose the top 8,000 distinct words as the vocabulary~\cite{CTR}. The average number of words per review is 32.9.

The statistics of the two datasets are summarized in Table~\ref{table:data}. As we can see, both datasets are very sparse and hence we hope improve performance by transferring knowledge from the auxiliary domain and exploiting the text content as well. Note that Amazon dataset are long text of product reviews (the number of average words per item is 32), while Cheetah Mobile is short text of news titles (the number of average words per item is 7).

\noindent
{\bf Evaluation Protocol } For item recommendation task, the leave-one-out (LOO) evaluation is widely used and we follow the protocol in~\cite{he2017neural}. That is, we reserve one interaction as the test item for each user. We determine hyper-parameters by randomly sampling another interaction per user as the validation/development set. We follow the common strategy which randomly samples 99 (negative) items that are not interacted by the user and then evaluate how well the recommender can rank the test item against these negative ones. Since we aim at top-K item recommendation, the typical evaluation metrics are hit ratio (HR), normalized discounted cumulative gain (NDCG), and mean reciprocal rank (MRR), where the ranked list is cut off at $topK=\{5,10,20\}$. HR intuitively measures whether the reserved test item is present on the top-K list, defined as: $HR = \frac{1}{|\mathcal{U}|} \sum\nolimits_{u \in \mathcal{U}} \delta(p_u \leq topK),$ where $p_u$ is the hit position for the test item of user $u$, and $\delta(\cdot)$ is the indicator function. NDCG and MRR also account for the rank of the hit position, respectively defined as: $NDCG = \frac{1}{|\mathcal{U}|} \sum\nolimits_{u \in \mathcal{U}} \frac{\log2} {\log(p_u + 1)}, \textrm{ and } MRR = \frac{1}{|\mathcal{U}|} \sum\nolimits_{u \in \mathcal{U}} \frac{1}{p_u}.$ A higher value with lower cutoff indicates better performance.

\begin{table*}[]
\centering
\caption{Comparison Results on the Amazon Dataset. The best baselines
are marked with asterisks and the best results are boldfaced.}
\label{table:results-amazon}
\resizebox{0.75\textwidth}{!}{	
\begin{tabular}{c | lll | lll | lll }
\hline \hline	
\multirow{2}{*}{Method}&\multicolumn{3}{c|}{$topK=5$}&\multicolumn{3}{c|}{$topK=10$}&\multicolumn{3}{c}{$topK=20$}\\
         & HR    & NDCG  & MRR   & HR    & NDCG  & MRR   & HR    & NDCG  & MRR \\
\hline 	
BPRMF    & .0810 & .0583 & .0509 & .1204 & .0710 & .0561 & .1821 & .0864 & .0602 \\
\hline	
CDCF     & .1295 & .0920 & .0797 & .2070 & .1167 & .0897 & .3841 & .1609 & .1015  \\
\hline	
CMF      & .1498 & .0950 & .0771 & .2224 & .1182 & .0863 & .3573 & .1521 & .0957  \\
\hline	
HFT      & .1077 & .0815 & .0729 & .1360 & .0907 & .0767 & .2782 & .1252 & .0854  \\
\hline	
TextBPR  & .1517 & .1208 & .1104 & .1777 & .1291 & .1138 & .2268 & .1414 & .1171 \\
\hline	
CDCF++   & .1314 & .0926 & .0800 & .2102 & .1177 & .0901 & .3822 & .1605 & .1016    \\
\hline	 	
MLP      & .2100 & .1486 & .1283 & .2836 & .1697 & .1371 & .3820 & .1899 & .1426 \\
\hline	
MLP++    & .2263 & .1626 & .1417 & .2992 & .1862 & .1514 & .3810 & .2069 & .1570 \\
\hline	
CSN      & .2340*& .1680*& .1462*& .3018*& .1898*& .1552*& .3944*& .2091*& .1605*\\
\hline		
LCMR     & .2024 & .1451 & .1263 & .2836 & .1678 & .1356 & .3951 & .1918 & .1420  \\
\hline	
TMH&{\bf .2575}&{\bf.1796}&{\bf.1550}&{\bf.3490}&{\bf.2077}&{\bf.1666}&{\bf.4443}&{\bf.2311}&{\bf.1727}\\
\hline
Improvement of TMH & 10.04\%& 6.90\%& 6.01\%& 15.63\%& 9.43\%& 7.34\%& 12.65\%& 10.52\%& 7.60\%\\
\hline
\end{tabular}
}
\end{table*}

\noindent
{\bf Baselines} We compare with various baselines, categorized as single/cross domain, shallow/deep, and hybrid methods.

\noindent
\resizebox{0.44\textwidth}{!}{
\begin{tabular}{|c|c|c|}
\hline
 Baselines          & Shallow  method   & Deep method \\
\hline
Single-Domain & BPRMF~\cite{BPRMF}             & MLP~\cite{he2017neural}   \\
\hline
Cross-Domain  & CDCF~\cite{CDCF}, CMF~\cite{CMF} & MLP++, CSN~\cite{CSN} \\
\hline
Hybrid  & HFT~\cite{HFT}, TextBPR~\cite{VBPR,TBPR} & LCMR~\cite{LCMR} \\
\hline
Cross + Hybrid  & CDCF++ & TMH (ours) \\
\hline
\end{tabular}
}

\begin{itemize}
\item {\bf BPRMF}, Bayesian personalized ranking~\cite{BPRMF}, is a latent factor model based on matrix factorization and pair-wise loss. It learns on the target domain only.
\item {\bf HFT}, Hidden Factors and hidden Topics~\cite{HFT}, adopts topic distributions to learn latent factors from text reviews. It is a hybrid method.
\item {\bf CDCF}, Cross-domain CF with factorization machines (FM)~\cite{CDCF}, is a cross-domain recommender which extends FM~\cite{FM}. It is a context-aware approach which applies factorization on the merged domains (aligned by the shared users). That is, the auxiliary domain is used as context. On the Mobile dataset, the context for a user in the target news domain is his/her history of app installations in the source app domain. The feature vector for the input is a sparse vector $\bm{x} \in \mathbb{R}^{m+n_T+n_S}$ where the non-zero entries are as follows: 1) the index for user id, 2) the index for target news id (target domain), and all indices for his/her installed apps (source domain).
\item {\bf CDCF++}: We extend the above CDCF model to exploit the text content. The feature vector for the input is a sparse vector $\bm{x} \in \mathbb{R}^{m+n_T+n_S+L}$ where the non-zero entries are augmented by the word features corresponding to the given user-item interaction. In this way, CDCF++ can learn from both the source domain and unstructured text information.
\item {\bf CMF}, Collective matrix factorization~\cite{CMF}, is a multi-relation learning approach which jointly factorizes matrices of individual domains. Here, the relation is the user-item interaction. On Mobile, the two matrices are $\bm{A}=$ ``user by news'' and $\bm{B}=$ ``user by app'' respectively. The shared user factors $\bm{P}$ enable knowledge transfer between two domains. Then CMF factorizes matrices $\bm{A}$ and $\bm{B}$ simultaneously by sharing the user latent factors: $\bm{A} \approx \bm{P}^T \bm{Q}_A$ and $\bm{B} \approx \bm{P}^T \bm{Q}_B$. It is a shallow model and jointly learns on two domains. CMF is a multi-objective shallow model for cross-domain recommendation. This can be thought of a non-deep transfer/multitask learning approach for cross-domain recommendation.
\item {\bf TextBPR} extends the basic BPRMF model by integrating text content. It computes the prediction scores by two parts: one is the standard latent factors, same with the BPRMF; and the other is the text factors learned from the text content. It has two implementations, the VBPR model~\cite{VBPR} and the TBPR model~\cite{TBPR} which are the same in essence.
\item {\bf MLP}, multilayer perceptron~\cite{he2017neural}, is a neural CF approach which learns the nonlinear interaction function using neural networks. It is a deep model learning on the target domain only.
\item {\bf MLP++}: We combine two MLPs by sharing the user embedding matrix, enabling the knowledge transfer between two domains through the shared users. It is a naive knowledge transfer approach applied for cross-domain recommendation.
\item {\bf CSN}, Cross-stitch network~\cite{CSN}, is a deep multitask learning model originally proposed for visual recognition tasks. We use the cross-stitch units to stitch two MLP networks. It learns a linear combination of activation maps from two networks and hence benefits from each other. Comparing with MLP++, CSN enables knowledge transfer also in the hidden layers besides the lower embedding matrices. CSN optimizes a multi-objective problem for cross-domain recommendation. This is a deep transfer learning approach for cross-domain recommendation.
\item {\bf LCMR}, Local and Centralized Memory Recommender~\cite{LCMR}, is a deep model for collaborative filtering with unstructured Text. The local memory module is similar to our MNet except that we only have one layer. LCMR is corresponding to the MNet component of our model. This is a deep hybrid method.
\end{itemize}

\noindent
{\bf Implementation } For BPRMF, we use LightFM's implementation\footnote{\url{https://github.com/lyst/lightfm}} which is a popular CF library. For CDCF and CDCF++, we adapt the official libFM implementation\footnote{\url{http://www.libfm.org}}. For CMF, we use a Python version reference to the original Matlab code\footnote{\url{http://www.cs.cmu.edu/~ajit/cmf/}}. For HFT and TextBPR, we use the code released by their authors\footnote{\url{http://cseweb.ucsd.edu/~jmcauley/}}. The word embeddings used in the TextBPR are pre-trained by GloVe~\cite{Pennington2014GloveGV}\footnote{\url{https://nlp.stanford.edu/projects/glove/}}. For latent factor models, we vary the number of factors from 10 to 100 with step size 10. For MLP, we use the code released by its authors\footnote{\url{https://github.com/hexiangnan/neural_collaborative_filtering}}. The MLP++ and CSN are implemented based on MLP. The LCMR model is similar to our MNet model and thus implemented in company. Our methods are implemented using TensorFlow. Parameters are randomly initialized from Gaussian $\mathcal{N}(0, 0.01^2)$. The optimizer is Adam with initial learning rate 0.001. The size of mini batch is 128. The ratio of negative sampling is 1. The MLP and MLP++ follows a tower pattern, halving the layer size for each successive higher layer. Specifically, the configuration of hidden layers in the base MLP network is $[64 \rightarrow 32 \rightarrow 16 \rightarrow 8]$ as reference in the original paper~\cite{he2017neural}. For CSN, it requires that the number of neurons in each hidden layer is the same and the configuration is $[64] * 4$ (equals $[64 \rightarrow 64 \rightarrow 64 \rightarrow 64]$). We investigate several typical configurations $\{16,32,64,80\} * 4$ . The dimension of embeddings is $d=75$.

\begin{table*}[]
\centering
\caption{Comparison Results on the Mobile Dataset. The best baselines
are marked with asterisks and the best results are boldfaced.}
\label{table:results-mobile}
\resizebox{0.75\textwidth}{!}{	
\begin{tabular}{c | lll | lll | lll }
\hline \hline	
\multirow{2}{*}{Method}&\multicolumn{3}{c|}{$topK=5$}&\multicolumn{3}{c|}{$topK=10$}&\multicolumn{3}{c}{$topK=20$}\\
         & HR    & NDCG  & MRR   & HR    & NDCG  & MRR   & HR    & NDCG  & MRR \\
\hline 	
BPRMF    & .4380 & .3971 & .3606 & .4941 & .4182 & .3694 & .5398 & .4316 & .3730 \\
\hline	
CDCF     & .5066 & .3734 & .3293 & .5325 & .4089 & .3441 & .5452 & .4374 & .3519  \\
\hline	
CMF      & .4789 & .3535 & .3119 & .5846 & .3879 & .3263 & .6662 & .4086 & .3320  \\
\hline	
HFT      & .4966 & .3617 & .3175 & .5580 & .4093 & .3365 & .6547 & .4379 & .3445  \\
\hline	
TextBPR  & .4948 & .4298 & .3826 & .5466 & .4499 & .3913 & .6123 & .4682 & .3958 \\
\hline	
CDCF++   & .4981 & .3693 & .3267 & .6055 & .4041 & .3411 & .6244 & .4335 & .3491    \\
\hline	 	
MLP      & .5380 & .4121 & .3702 & .6176 & .4381 & .3810 & .6793 & .4529 & .3851 \\
\hline	
MLP++    & .5524 & .4284 & .3871 & .6319 & .4535 & .3976 & .6910 & .4691 & .4019 \\
\hline	
CSN      & .5551*& .4323*& .3920*& .6327*& .4574*& .4025*& .6908 & .4732*& .4068*\\
\hline		
LCMR     & .5476 & .4189 & .3762 & .6311 & .4460 & .3874 & .6927*& .4619 & .3918 \\
\hline	
TMH    &{\bf .5664} & {\bf .4427} & {\bf .4018} & {\bf .6438} & {\bf .4680} & {\bf .4124} & {\bf .6983} & {\bf .4820} & {\bf .4163}\\
\hline
Improvement of TMH & 2.04\% & 2.42\% & 2.51\% & 1.75\% & 2.32\% & 2.47\% & 0.81\% & 1.86\% & 2.34\% \\
\hline
\end{tabular}
}
\end{table*}

\subsection{Comparison Results}\label{exp:results}											
In this section, we report the recommendation performance of different methods and discuss the findings. The comparison results are shown in Table~\ref{table:results-mobile} and Table~\ref{table:results-amazon} respectively on the Mobile and Amazon datasets where the last row is the relative improvement of ours vs the best baseline. We have the following observations. Firstly, we can see that our proposed neural models are better than all baselines on the two datasets at each setting, including the base MLP network, shallow cross-domain models (CMF and CDCF), deep cross-domain models (MLP++ and CSN), and hybrid methods (HFT and TextBPR, LCMR). These results demonstrate the effectiveness of the proposed neural model.

On the Mobile dataset, the differences between TMH and other methods are more pronounced for small numbers of recommended items including top-5 or top-10 where we achieve average 2.25\% relative improvements over the best baseline. This is a desirable feature since we often recommend only a small number of top ranked items to consumers to alleviate the information overload issue.

Note that the relative improvement of the proposed model vs. the best baseline is more significant on the Amazon dataset than that on the Mobile dataset, obtaining average 9.56\% relative improvements over the best CSN baseline, though the Amazon is sparser than the Mobile (see Table~\ref{table:data}). We show the benefit of combining text content by comparing with CSN. One explanation is that the relatedness of the Men and Sports domains is closer than that between the news and app domains. This will benefit all cross-domain methods including CMF, CDCF, MLP++, and CSN, since they exploit information from both two domains. Another explanation is that the text content contains richer information on the Amazon dataset. As it is shown in Table~\ref{table:data}, the average words in the product reviews are longer that in the news titles. This will benefit all hybrid methods including HFT, TextBPR, and LCMR. We show the benefit of transferring source items by comparing with LCMR.

The hybrid TextBPR model composes a document representation by averaging the words's embeddings. This can not distinguish the important words to match the user preferences. This may explain that it has difficulty in improving the recommendation performance when integrating text content. For example, it can not consistently outperform the pure CF method, MLP. The cross-domain CSN model transfers every representations from the source network with the same coefficient. This may have a risk in transferring the noise and harm the performance, as pointed out in its sparse variant~\cite{CoNet}. On the Amazon dataset, it loses to the proposed model by a large margin (though TMH leverages content information). In contrast, the memory and transfer components are both selective to extract useful information based on the attention mechanism. This may explain that our model is consistently the best at all settings.

There is a possibility that the noise from auxiliary domain and some irrelevance information contained in the unstructured text propose a challenge for exploiting them. This shows that the proposed model is more effective since it can select useful representations from the source network and attentively focus on the important words to match preferences of users.

In summary, the empirical comparison results demonstrate the superiority of the proposed neural model to exploit the text content and source domain knowledge for recommendation.

\subsection{Impact of Unstructured Text and Auxiliary Domain}\label{exp:impact}

\begin{figure}
\centering
\includegraphics[width=4.0cm, height=3.5cm]{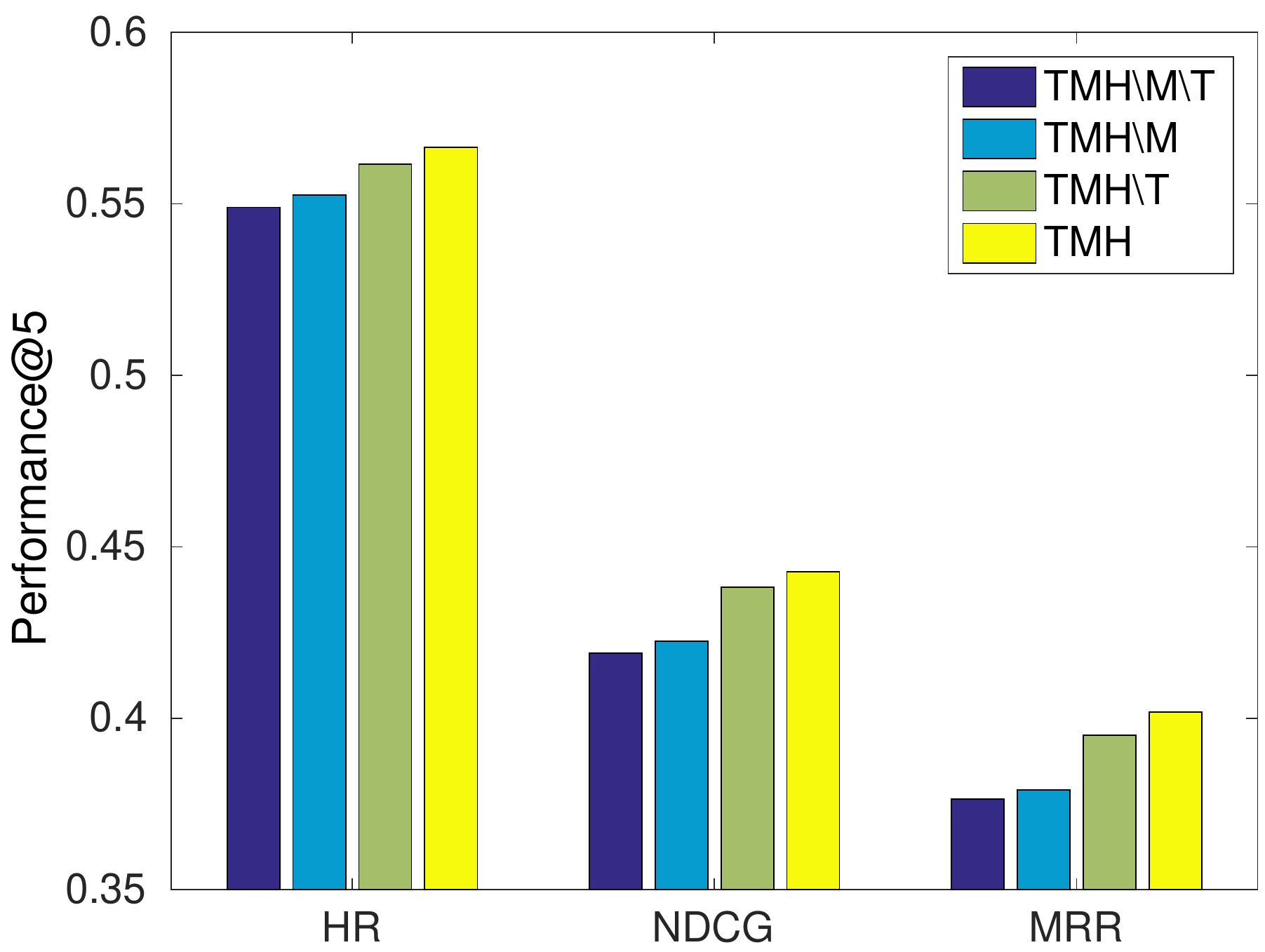}\label{fig:impact5-mobile}
\includegraphics[width=4.0cm, height=3.5cm]{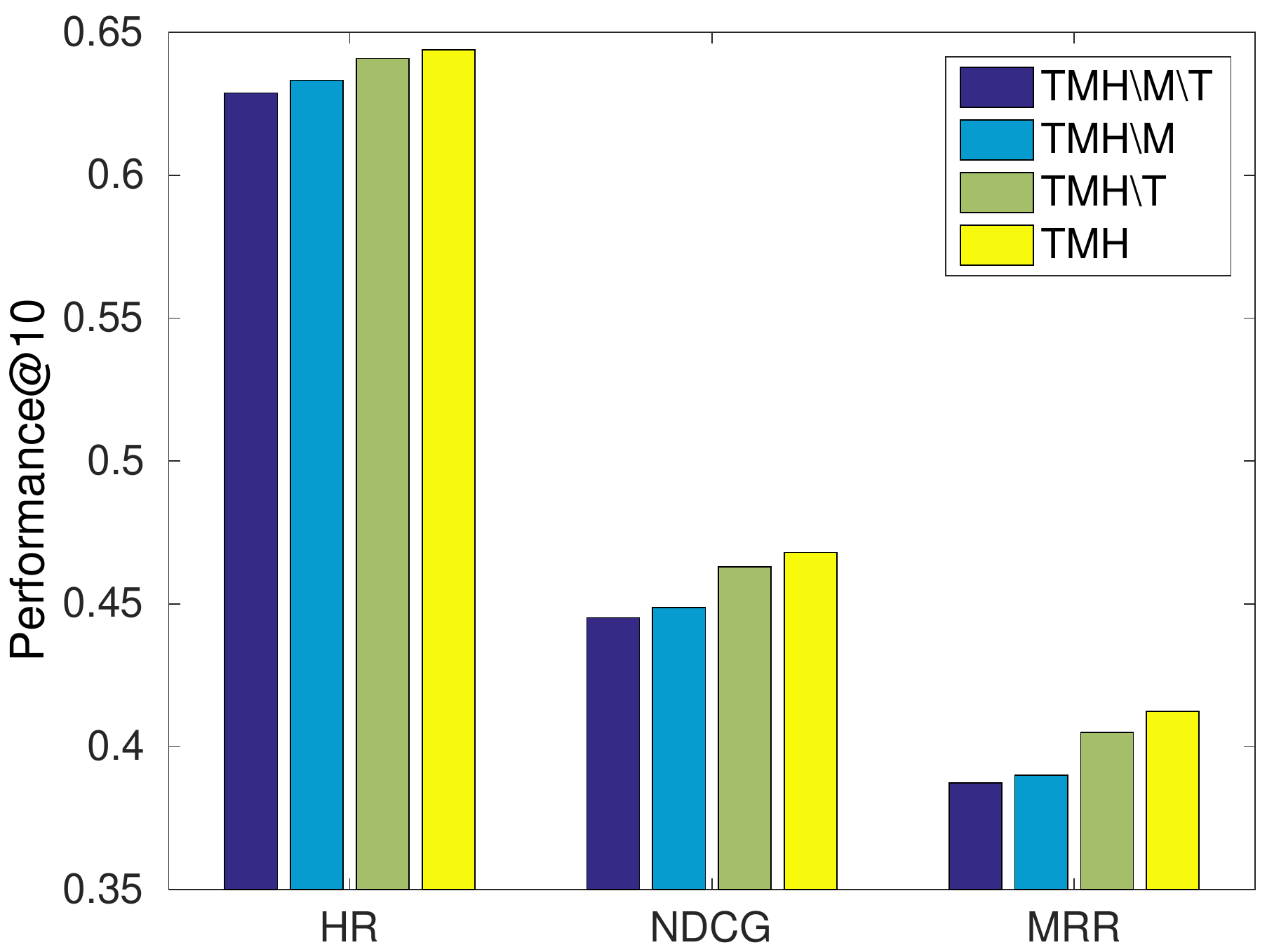}\label{fig:impact10-mobile}
\includegraphics[width=4.0cm, height=3.5cm]{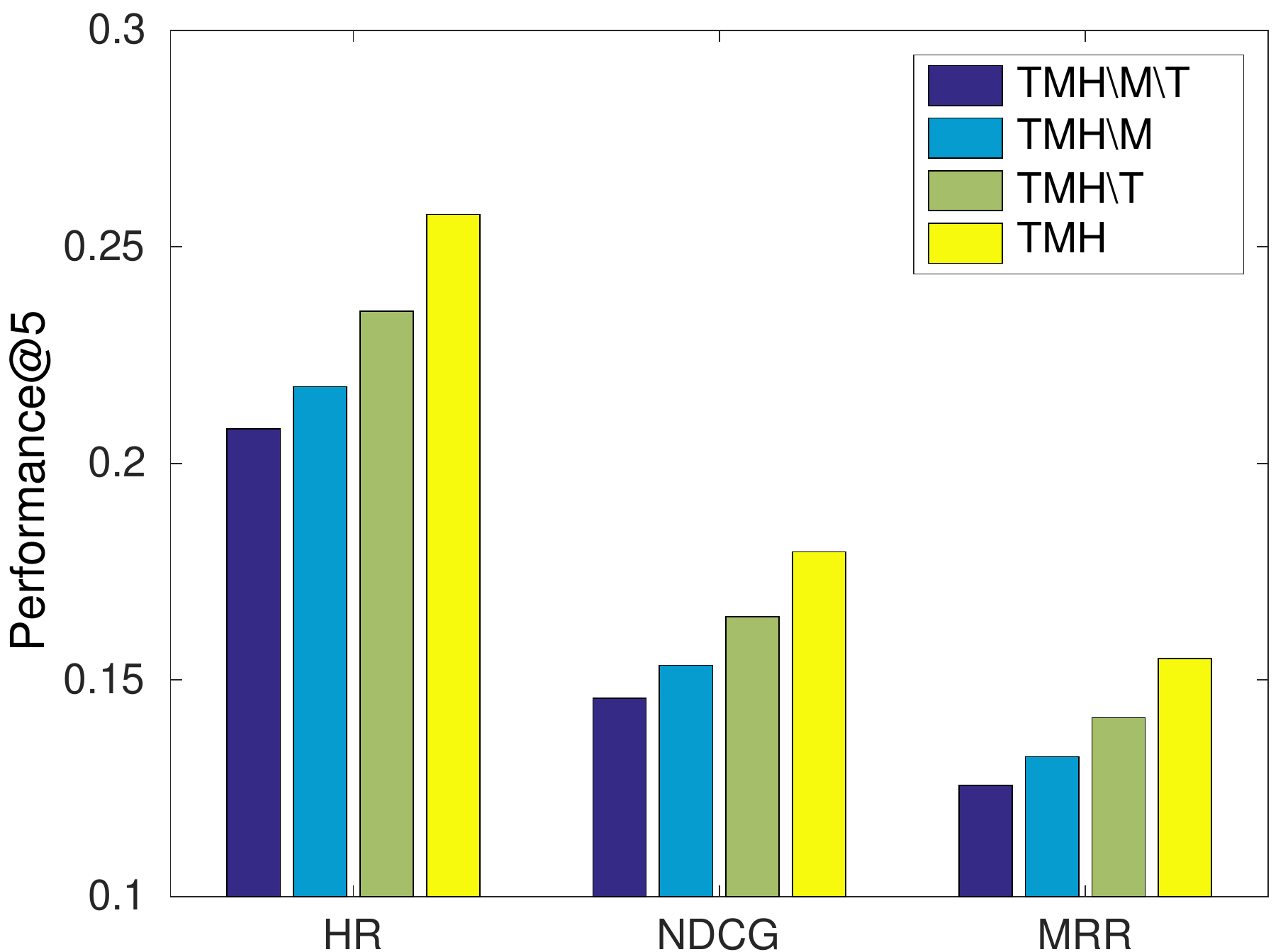}\label{fig:impact5-amazon}
\includegraphics[width=4.0cm, height=3.5cm]{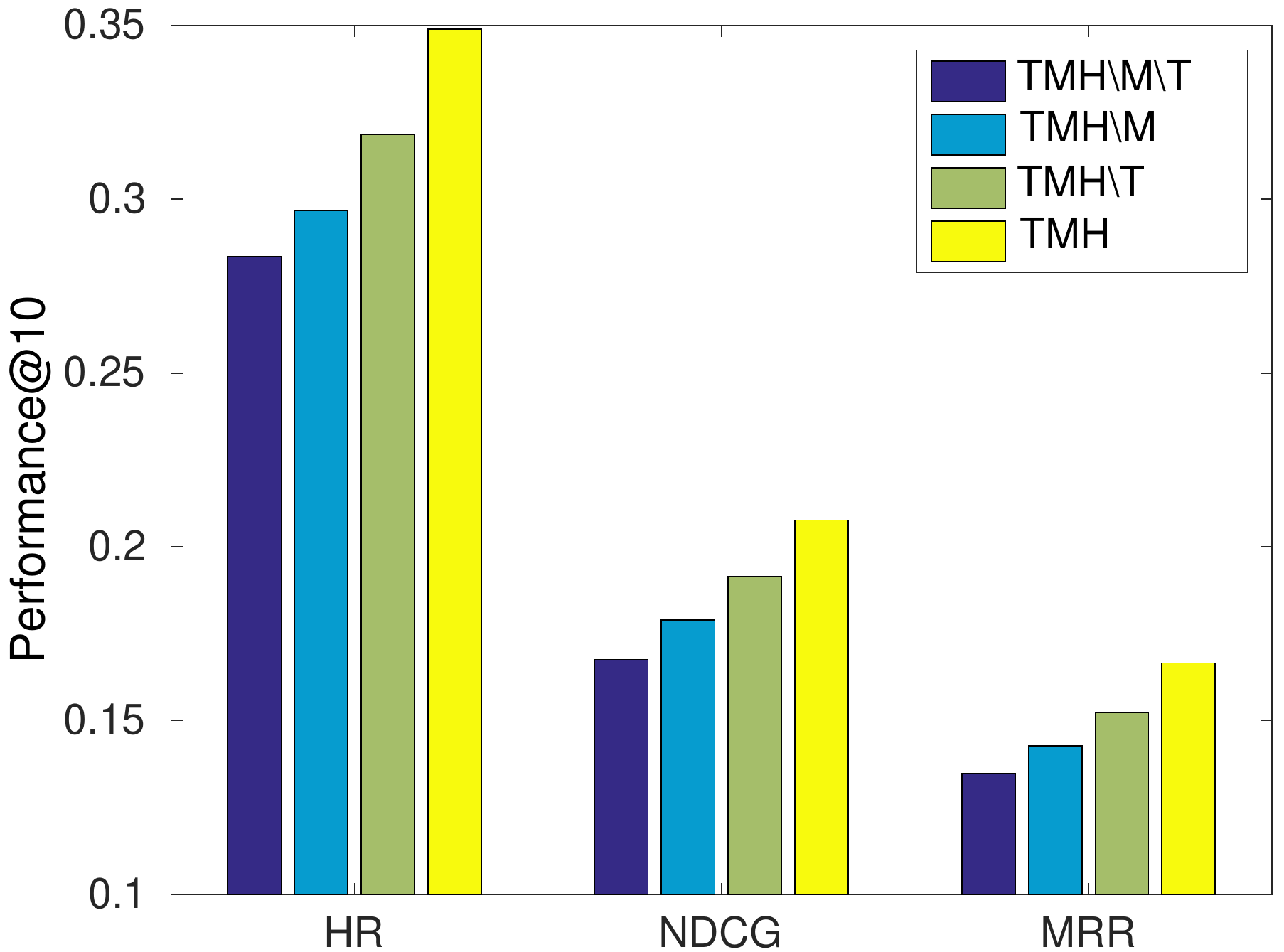}\label{fig:impact10-amazon}
    \caption{Contributions from Unstructured Text (MNet) and Cross-Domain Knowledge (TNet) on the Mobile (Top) and Amazon (Bottom) Datasets.}
\label{fig:components}
\end{figure}

We have shown the effectiveness of the two memory and transfer components together in the proposed framework. We now investigate the contribution of each network to the TMH by eliminating the impact of text content and source domain from it in turn:
\begin{itemize}
\item {\bf TMH$\backslash$M$\backslash$T}: Eliminating the impact of both content and source information from TMH. This is a collaborative filtering recommender. Actually, it is equivalent to a single hidden layer MLP model.
\item {\bf TMH$\backslash$M}: Eliminating the impact of content information (MNet) from TMH. This is a novel cross-domain recommender which can adaptively select source items to transfer via the attentive weights.
\item {\bf TMH$\backslash$T}: Eliminating the impact of source information (TNet) from TMH.  This is a novel hybrid filtering recommender which can attentively match word semantics with user preferences.
\end{itemize}

The ablation analyses of TMH and its components are shown in Figure~\ref{fig:components}. The performance degrades when either memory or transfer modules are eliminated. This is understandable since we lose some information. In other words, the two components can extract useful knowledge to improve the recommendation performance. For example, TMH$\backslash$T and TMH$\backslash$M respectively reduce 1.1\% and 4.3\% relative NDCG@10 performance by comparing with TMH on the Mobile dataset (they are 8.5\% and 16.1\% on Amazon), suggesting that both memory and transfer networks learn essential knowledge for recommendation. On the evaluated two datasets, removing the memory component degrades performance worse than that of removing the transfer component. This may be due to that the text content contains richer information or the source domain contains much more noise or both.

\subsection{Improvement on Cold Users and Items}\label{exp:cold}

The cold-user and cold-item problems are common issues in recommender systems. When new users enter into a system, they have no history that can be exploited by the recommender system to learn their preferences, leading to the cold-user start problem. Similarly, when latest news are released on the Google News, there are no reading records that can be exploited by the recommender system to learn users' preferences on them, leading to the cold-item start problem. In general, it is very hard to train a reliable recommender system and make predictions for users and items that have few interactions. Intuitively, the proposed model can alleviate both the cold-user and cold-item start issues. TMH alleviates the cold-user start issue in the target domain by transferring his/her history from the related source domain. TMH alleviates the cold-item start issue by exploiting the associated text content to reveal its properties, semantics, and topics. We now investigate that TMH indeed improves the performance over the cold users and items by comparing with the pure neural collaborative filtering method, MLP.

We analyse the distribution of missed hit users (MHUs) of TMH and MLP (at cutoff 10). We expect that the cold users in MHUs of MLP can be reduced by using the TMH model. The more amount we can reduce, the more effective that TMH can alleviate the cold-user start issues. The results are shown in Figure~\ref{fig:cold-users} where the number of training examples can measure the ``coldness'' of a user. Naturally, the MHUs are most of the cold users who have few training examples. As we can see, the number of cold users in MHUs of MLP is higher than that of TMH. If the cold users are defined as those with less than seven training examples, then TMH reduces the number of cold users from 4,218 to 3,746 on the Amazon dataset, achieving relative 12.1\% reduction. On the Mobile dataset, if the cold users are those with less than
ten training examples (Mobile is denser than Amazon), then TMH reduces the number of cold users from 1,385 to 1,145 on the Mobile dataset, achieving relative 20.9\% reduction. These results show that the proposed model is effective in alleviating the cold-user start issue. The results on cold items are similar and we omit them due to the page limit.

\begin{figure}
\centering
\includegraphics[width=4.1cm,height=3.6cm] {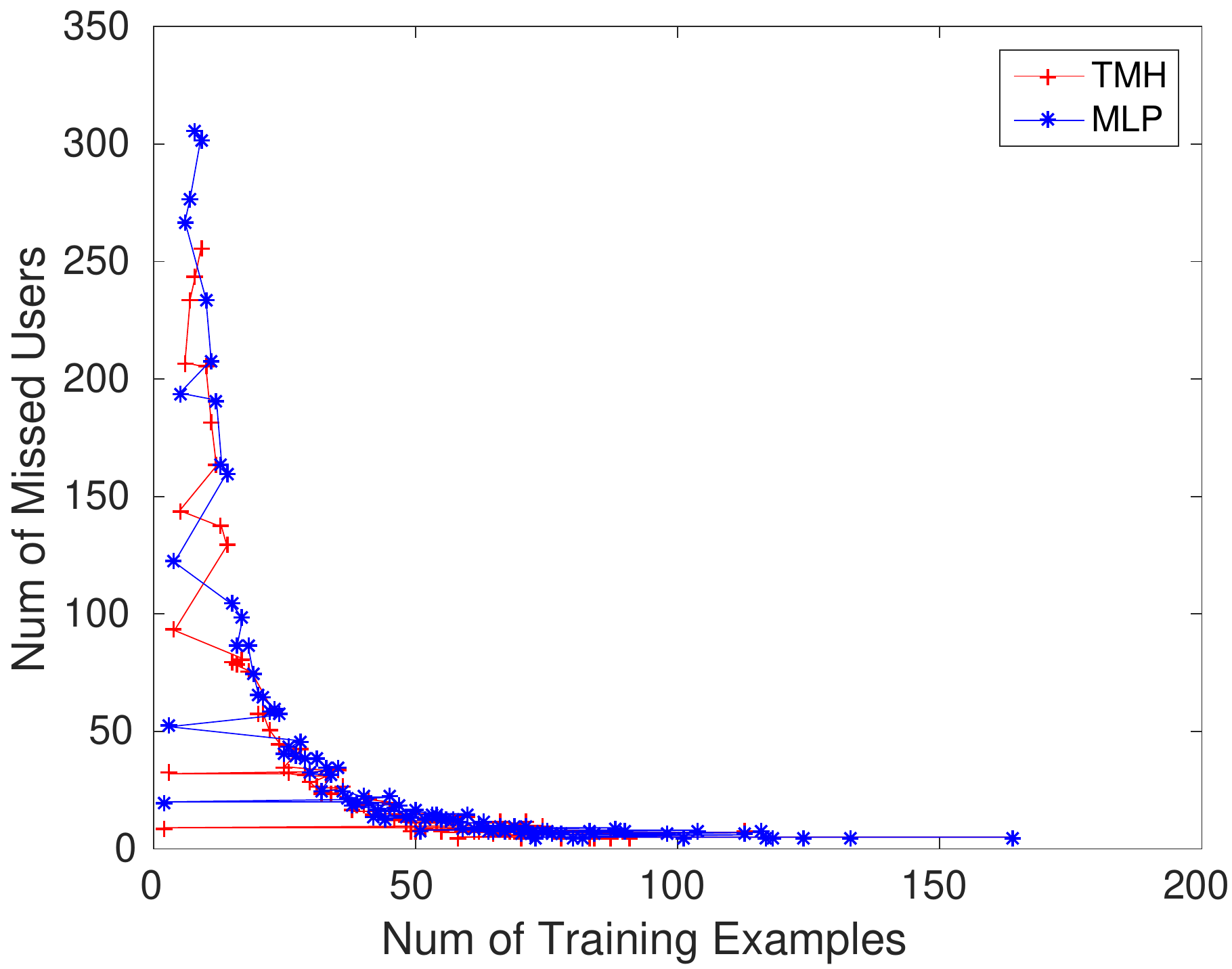}\label{fig:cold-users-mobile}
\includegraphics[width=4.1cm,height=3.6cm] {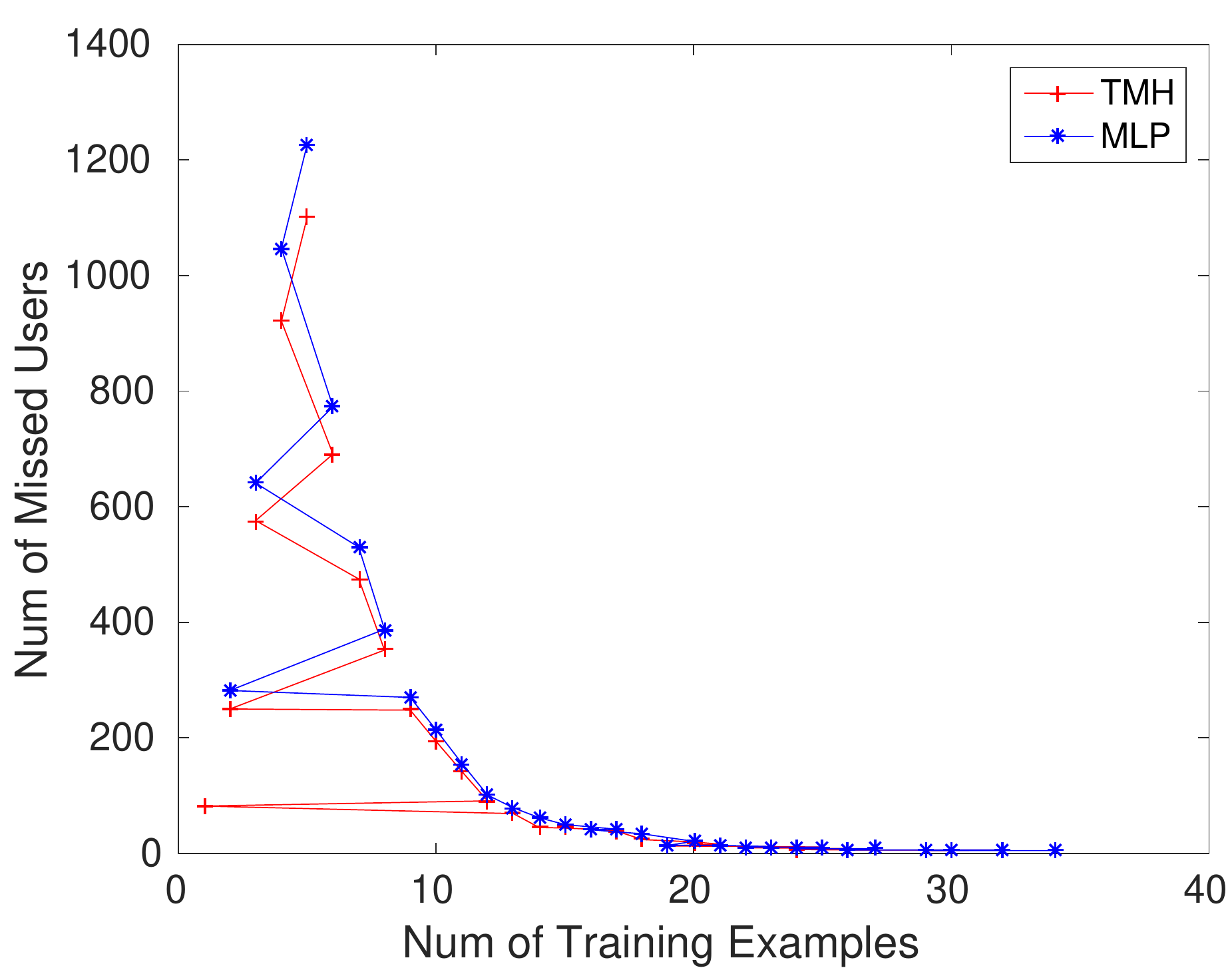}\label{fig:cold-users-amazon}
\caption{The Missed Hit Users Distribution (not normalized) Over the Number of Training Examples on the Mobile (Left) and Amazon (Right) Datasets.}
\label{fig:cold-users}
\end{figure}

\section{Conclusion}\label{paper:conclusion}

It is shown that the text content and the source domain knowledge can help improve recommendation performance and can be effectively integrated under a neural architecture. The sparse target user-item interaction matrix can be reconstructed with the knowledge guidance from both of the two kinds of information, alleviating the data sparse issue. We proposed a novel deep neural model, TMH, for cross-domain recommendation with unstructured text. TMH smoothly enables transfer meeting hybrid. TMH consists of a memory component which can attentively focus important words to match user preferences and a transfer component which can selectively transfer useful source items to benefit the target domain. These are achieved by the attentive weights learned automatically. TMH shows better performance than various baselines on two real-world datasets under different settings. The results demonstrate that our combine model outperforms the baseline that relies only on memory networks (LCMR~\cite{LCMR}) and outperforms the baseline that relies only on the transfer networks (CSN~\cite{CSN}). Additionally, we conducted ablation analyses to understand contributions from the two memory and transfer components, showing the necessity to combine transfer and hybrid. We quantify the amount of missed hit cold users (and items) that TMH can reduce by comparing with the pure CF method, showing that TMH is able to alleviate the cold-start issue.

In real world services, data sources may belong to different providers (e.g. product reviews provided by Amazon while social relations provided by Facebook). The data privacy is a big issue when we combine the multiple data sources. In future work, it is worth developing new learning techniques to learn a combined model while protecting user privacy.

\bibliographystyle{ACM-Reference-Format}
\bibliography{sample-bibliography}

\end{document}